\documentclass{article}
\usepackage{spconf,amsmath,graphicx}

\title{Effective Text Adaptation for LLM-based ASR through Soft Prompt Fine-tuning}
\name{Yingyi Ma, Zhe Liu, Ozlem Kalinli}
\address{Meta AI, Menlo Park, CA, USA}
\input{./Definitions}
\usepackage{subcaption}
\usepackage{multirow}
\usepackage{graphics}

\begin{document}
%
\maketitle
\begin{abstract}
The advent of Large Language Models (LLM) has reformed the Automatic Speech Recognition (ASR). Prompting LLM with audio embeddings to generate transcriptions becomes the new state-of-the-art ASR. Despite LLMs being trained with an extensive amount of text corpora, high-quality domain-specific text data can still significantly enhance ASR performance on domain adaptation tasks. Although LLM-based ASR can naturally incorporate more text corpora by fine-tuning the LLM decoder, fine-tuning such ASR on text-only data without paired prompts may diminish the effectiveness of domain-specific knowledge. To mitigate this issue, we propose a two-step soft prompt fine-tuning strategy that enhances domain-specific text adaptation. Experimental results show that text adaptation with our proposed method achieved a relative up to 9\% Word Error Rate (WER) reduction and up to 18\% Entity Error Rate (EER) reduction on the target domain compared to the baseline ASR. Combining this with domain-specific Language Model (LM) fusion can further improve the EER by a relative 2-5\%.
\end{abstract}

\section{Introduction}

In recent years, Large Language Models (LLMs) have made significant strides in various natural language tasks \cite{AchAdlAga2023, AniDaiFir2023, TouLavIza2023}. This substantial improvement in text-based tasks has inspired numerous studies to extend the capabilities of LLMs to perceive non-textual information, thereby bridging the gap between various modalities \cite{LyuWuWan2023, HanZhaSha2023, WuFeiQu2023}. To facilitate the quick adaptation of LLMs to new tasks, research has shown that prompt design is highly effective. Inspired by this, recent studies have integrated speech with large language models for Automatic Speech Recognition (ASR) tasks, where transcription generation is conditioned on acoustic information as prompts \cite{FatWuLak2023, WuGauChe2023}. This LLM-based ASR approach employs a pretrained audio encoder to encode and project audio features into the text embedding space. The encoded audio embeddings are then prepended to the text embeddings to serve as prompts. Leveraging the exceptional generation capability, LLM-based ASR has demonstrated superior performance over conventional ASR on multiple tasks.

 Bountiful text corpora have been used in training LLMs, thereby enabling LLM-based ASR to perform proficiently on general ASR tasks. However, the importance of high-quality, domain-specific data, particularly from entity-heavy domains, cannot be understated for adaptation tasks. Given the relative ease of access to text-only data compared to audio-text paired data, a multitude of studies have embarked on exploring the implementation of text adaptation on conventional ASR models through fine-tuning the model decoder \cite{PylUkkKil2021, MenGauKan2021, CheMenPar2022, MenChePra2023, GaoCheYan2021}. One distinct advantage of LLM-based ASR is that the decoder component functions as an actual language model, thus naturally accommodating text adaptation through decoder fine-tuning. However, recent studies have indicated that improperly managed prompts can have a detrimental impact on task adaptation \cite{WanSiLi}. The absence of audio features can lead to a prompt mismatch between training and inference, potentially undermining the effectiveness of domain knowledge adaptation. In terms of implicit prompt handling, strategies such as prompt tuning \cite{LesAlCon2021} and prefix tuning \cite{LiLia2021} have been proposed to facilitate efficient task adaptation.

Inspired by previous studies, we propose a novel approach for domain adaptation fine-tuning, which involves the learning of pseudo audio embeddings as prompts. Our approach is characterized by a two-step soft prompt fine-tuning process. In the first step, a trainable soft prompt is learned as a domain-specific pseudo audio embedding. In the second step, the text adaptation fine-tuning is conducted by freezing the learned prompt.
In addition to this, we have explored other strategies for prompt handling and adaptation approaches like external LM fusion. These explorations aim to close the gap for text adaptation of LLM-based ASR. We have demonstrated the effectiveness of our proposed method on two entity-heavy domain adaptation tasks. The results indicate a significant improvement in domain entity recognition. Furthermore, this improvement is found to be extendable when our method is combined with external LM fusion.

\begin{figure*}[t!]
    \centering
    \begin{subfigure}[t]{0.32\textwidth}
        \centering
        \includegraphics[height=1.2in]{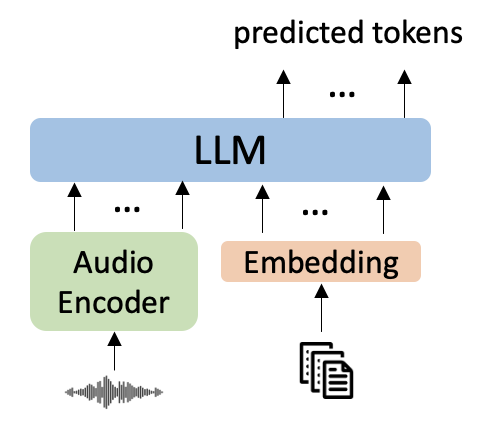}
        \caption{LLM-based ASR}
    \end{subfigure}
    ~ 
    \begin{subfigure}[t]{0.32\textwidth}
        \centering
        \includegraphics[height=1.2in]{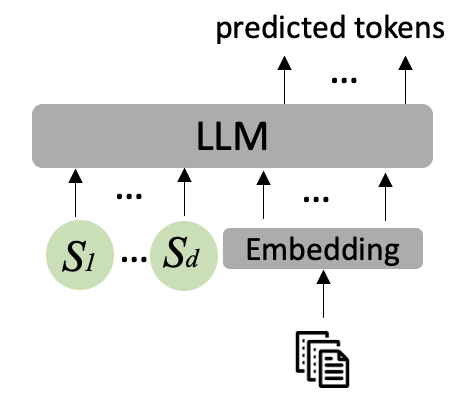}
        \caption{train soft prompt}
    \end{subfigure}
    ~
    \begin{subfigure}[t]{0.32\textwidth}
        \centering
        \includegraphics[height=1.2in]{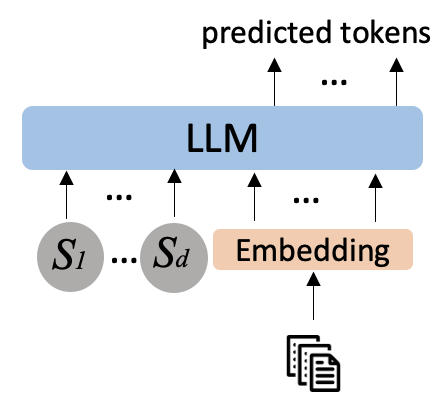}
        \caption{fine-tune LLM with soft prompt}
    \end{subfigure}
    \caption{The training process of LLM-based ASR is illustrated in (a). The domain adaptation fine-tuning with soft prompt as pseudo audio embedding is illustrated in (b) and (c). We first train soft prompt $S\in \RR^{d \times e}$ by freezing all other components, then fine-tune LLM with the trained soft prompt for more effective text adaptation. The freezed components are shown in grey.}
    \label{fig:training_process}
\end{figure*}

\section{Related Works}

LLM-based ASR has been highlighted in several recent studies \cite{WuGauChe2023, FatWuLak2023, CheLiHu2024}. These studies typically encode audio features into the text embedding space, then prompt the LLM with audio input to generate corresponding text. To address the modality gap caused by the high sampling rate of acoustic data, \cite{WuGauChe2023} proposed to leverage a CTC compressor which is trained to select representative frames, \cite{FatWuLak2023} suggested stacking frames within a certain interval to represent a longer duration.
Despite these advancements, all the aforementioned methods require training with a large amount of audio-text paired data and primarily focus on general domain ASR tasks. The adaptation of ASR to new domains without paired data within the new framework of LLM-based ASR remains an unexplored area.

In terms of text injection and adaptation, numerous explorations have been conducted on top of conventional end-to-end ASR models like RNN-T \cite{PylUkkKil2021, CheMenPar2022, MenChePra2023, BapChuWu2021,CheZhaRos2022a}. These studies include injecting text through modality matching approaches \cite{BapChuWu2021,CheZhaRos2022a,tanGonDon2022,ThoKuoKin2022,SaiPraBap2023}, and leveraging Text-to-Speech (TTS) techniques to generate paired audio features \cite{BatKorSha2023, UenKaw2022, KarWatIwa2019, HorAstHay2019, CheZhaRos2022b}.
A common approach towards modality matching involves leveraging duration models, typically defined as an up-sampling operator. Duration models up-sample text embeddings to coordinate with frames distribution through a pre-trained alignment model \cite{ThoKuoKin2022, SaiPraBap2023}. The up-sampled text embeddings are then randomly masked and treated as pseudo audio embeddings.
TTS related approaches have also shown promising results in domain adaptation tasks. However, the efficiency of these methods is often questioned due to the high computational cost of generating audio features. Moreover, unreliable audio features can introduce discordant noise, leading to negative effects in adaptation training and making the performance of such approaches heavily dependent on the quality of TTS models \cite{KyrSaoKin2021, HuArmShr2022}. This drawback is particularly significant in entity-heavy domains.

Towards guiding pre-trained LLMs to solve domain-specific tasks, numerous in-context prompting works have been established \cite{LesAlCon2021, ZhaLiChe2021, LiLia2021, WanSiLi}. Beyond full fine-tuning, parameter-efficient fine-tuning strategies like prompt-tuning \cite{LesAlCon2021} or prefix-tuning \cite{LiLia2021} have demonstrated effective results. These studies adapt the pre-trained model to specific tasks with only a small number of tunable parameters in prompts.
However, the primary goal of these works is to achieve fast adaptation of LLMs with lite fine-tuning. The learned prompt or prefix is not readily interpretable, nor their potential for further improving domain adaptation tasks is explored.

\section{Text Adaptation of LLM-based ASR}

A typical LLM-based ASR model is introduced in the work of \cite{FatWuLak2023}. This model is composed of three integral components: Audio encoder $\Acal$ which transforms audio features to audio embeddings, Text embedding $\Ecal$ which maps text features to embeddings, Decoder $\Dcal$ which is typically designed with an autoregressive LLM architecture. All components can be initialized from pre-trained modules. For a source domain $\tau$ with audio text paired data corpus $C =\{ (\ab_s, \xb_s) \}$, the LLM-based ASR model is trained by leveraging audio features as prompts at embedding space. The audio embedding prompts can be obtained through forwarding audio feature to audio encoder $\Acal(\ab_s)$. The training goal is then to maximize the likelihood of $\xb_s$ conditioned on audio embedding prompts:
\begin{align}
\argmax_{\Acal, \Ecal, \Dcal} \sum_{(\ab_s, \xb_s) \in C}P_{\Ecal, \Dcal}(\xb_s|\Acal(\ab_s))
\end{align}

We then consider a domain adaptation task, where the target domain $\zeta$ is different from the source domain $\tau$ and we only possess text records $Q =\{ (\xb_t) \} \in \zeta$ but have no access to text audio paired data. The goal is then to adapt text data to the LLM-based ASR model in an effective manner. Different from conventional ASR models, the LLM decoder architecture is inherently capable of accommodating text-only data. To fine-tune the decoder $\Dcal$ (and embedding $\Ecal$) of such a model, we can consider the following strategies:

\begin{itemize}
\item Pre fine-tune:  We fine-tune the LLM on $Q$ prior to training the ASR on $C$. The ASR model is then initialized with the text-adapted decoder, thereby increasing the likelihood for text from the target domain. This approach is straightforward for fine-tuning the decoder, as it occurs before the training conditioned on the audio embedding prompts. Consequently, there is no need to specifically manage the prompt when fine-tuning on $Q \in \zeta$. However, the early stage fine-tuning may be susceptible to catastrophic forgetting once the ASR is further fine-tuned on $C \in \tau$. 

\item Post fine-tune: To ensure the model retains the knowledge from the target domain, we can fine-tune the decoder of the ASR on $Q$ after it is trained on $C$. This strategy offers more flexibility as we can fine-tune the decoder only when domain adaptation is required. The challenge is that ASR was trained using audio embedding as prompts and $Q$ lacks audio features. The most straightforward method is to fine-tune the decoder with no prompt or an empty prompt. However, this mismatch in conditions may reduce the effectiveness of incorporating target domain knowledge.
\end{itemize}

\section{Pseudo Audio Embeddings}

Several studies have investigated the construction of pseudo audio inputs to incorporate text into traditional ASR models. Prior research \cite{SaiPraBap2023, ThoKuoKin2022} has demonstrated the significance of upsampling text embeddings to align with speech modality representations. To increase the complexity of the task for more effective training, these upsampled text embeddings were randomly masked.
In the training of LLM-based ASR, audio embeddings are also encoded based on a fixed speech duration, which simplifies the transfer of intuition. However, training with masked text sequences could be an over simple task for fine-tuning LLM decoders like the LLaMA, potentially leading to ineffective knowledge injection.

\subsection{Soft Prompt as Pseudo Audio Embedding}
\label{sec:soft_prompt}
Rather than explicitly aligning a pseudo audio embedding through upsampling and masking, we can alternatively seek to identify an implicit domain-specific pseudo audio embedding that can be shared across the entire domain. To achieve this, we propose a fine-tuning strategy that can be divided into two steps:

\subsubsection{domain specific soft prompt tuning}
Before fine-tuning the decoder on the target domain corpus $Q \in \zeta$, we first learn a trainable soft prompt situated in the audio embedding space. We denote this soft prompt as $S_{\zeta} \in \RR^{d \times e}$, where $d$ and $e$ represent the length of the soft prompt and the embedding dimension respectively. We train $S_{\zeta}$ on corpus $Q$ by freezing all parameters in the LLM-based ASR:
\begin{align}
\argmax_{S_{\zeta}} \sum_{\xb_t \in Q}P_{\Ecal, \Dcal}(\xb_t|S_{\zeta})
\end{align}
As shown in Figure \ref{fig:training_process}(b), during the training phase, we apply $S_{\zeta}$ as prompt embeddings. The audio encoder $\Acal$ is not employed in this stage due to the absence of provided audio features. 
Given that the training of the LLM-based ASR is conditioned on audio embeddings as prompts, optimizing the above objective allows $S_{\zeta}$ to learn domain-specific audio representations. Thereby, $S_{\zeta}$ can serve as pseudo audio embeddings to guide the text injection in the subsequent step.

\subsubsection{fine-tune decoder with soft prompt}

The soft prompt $S_{\zeta}$, learned in the previous step, should be capable of mimicking the audio embedding prompt from the target domain. Subsequently, as shown in Figure \ref{fig:training_process}(c), we fine-tune the decoder $\Dcal$ on $Q$ by supplying $S_{\zeta}$ as the prompt during training. Optimizing the decoder conditioned on such a trained pseudo prompt can result in reduced condition mismatch at the time of inference.

\subsection{Inference}
Once the decoder is fine-tuned on $Q$, the duty of soft prompt is done. At the inference time, the soft prompt is no longer utilized as we have access to the actual audio feature. Given that the decoder has been tuned under the guidance of the domain-specific pseudo prompt, it becomes easier for the model to recall the knowledge acquired from the target domain.

\subsection{Discussion}

Given that text sentences from different domains may vary in length, it is crucial that the fine-tuned soft prompts align with the preferred length of each domain to optimize text adaptation performance. The most direct method is to adjust the length of soft prompts as hyperparameters. Additionally, one could determine the embedding length of the centroid of the text sets and then set the soft prompt length using the duration handling approaches described in \cite{SaiPraBap2023}.
Given the abundant availability of text corpora in the real world, text adaptation has the potential to benefit multiple domains simultaneously or could specifically be applied to enhance general ASR performance. In such scenarios, the soft prompt can be flexibly configured either corpus-wise or task-wise, allowing for tailored adjustments that meet the specific needs of each application.

\section{Experiments}

\subsection{Datasets}

The statistics of data splits across each domain are summarized in Table \ref{table:data_summary}. Our experiments were conducted on in-house collected datasets, given the scarcity of public speech datasets available for entity-heavy domain adaptation tasks.
The \textit{video} corpus is sourced from public social media platforms. For the \textit{music} and \textit{chatbot} domains, the test splits are produced from a third-party data supplier. \textit{music} domain contains voice commands of music entities, e.g. \texttt{play <song name>}. Three test sets are collected for \textit{music} domain based on different album, artist, song name popularities. \textit{chatbot} domain contains arbitrary questions one may be interested in asking a LLM, e.g. \texttt{Which is the best restaurant in New York city}. Based on the focus of different entities, we split the test set into three subsets, namely organization (\textit{org.}), technology (\textit{tech.}), and \textit{food}. All these splits undergo a de-identification process before transcription. 
For text-only splits, we utilize in-house collected domain-specific entities as seed data. We then employ the LLaMA2-Chat 13B model to generate text surrounding these entities. 

\begin{table}[t!]
\caption{Summary of data splits from multiple domains}
\label{table:data_summary}
\begin{tabular}{l|ccc}
\hline
Domain                   & \multicolumn{3}{c}{Splits}                                                                                                                                                                                                                                                              \\ \hline
                         & \multicolumn{1}{l}{\begin{tabular}[c]{@{}l@{}}Text-only\\ (\# text records)\end{tabular}} & \multicolumn{1}{l}{\begin{tabular}[c]{@{}l@{}}Text-audio\\ (\# utts)\end{tabular}} & \multicolumn{1}{l}{\begin{tabular}[c]{@{}l@{}}Test\\ (\# utts)\end{tabular}}                           \\ \hline
\textit{video}                    & -                                                                                         & 50M                                                                                & -                                                                                                      \\ \hline
\multirow{3}{*}{\textit{music}}   & \multirow{3}{*}{3M}                                                                       & \multirow{3}{*}{-}                                                                 & \multirow{3}{*}{\begin{tabular}[c]{@{}c@{}}\textit{music 1}:  1.1k\\ \textit{music 2}:  1.4k\\ \textit{music 3}:  3.1k\end{tabular}} \\
                         &                                                                                           &                                                                                    &                                                                                                        \\
                         &                                                                                           &                                                                                    &                                                                                                        \\ \hline
\multirow{3}{*}{\textit{chatbot}} & \multirow{3}{*}{10M}                                                                      & \multirow{3}{*}{-}                                                                 & \multirow{3}{*}{\begin{tabular}[c]{@{}c@{}}\textit{food}:    0.4k\\ \textit{org.}:     0.5k\\ \textit{tech.}:    0.2k\end{tabular}}  \\
                         &                                                                                           &                                                                                    &                                                                                                        \\
                         &                                                                                           &                                                                                    &                                                                                                        \\ \hline
\end{tabular}

\bigskip

\caption{Text adaptation performance with no prompt}
\label{table:pre_finetune}
\resizebox{\columnwidth}{!}{
\begin{tabular}{l|ll|ll|ll}
\hline
\multicolumn{1}{c|}{\multirow{2}{*}{\begin{tabular}[c]{@{}c@{}}Text \\ adaptation\end{tabular}}} & \multicolumn{2}{c|}{\textit{music 1}}    & \multicolumn{2}{c|}{\textit{music 2}}    & \multicolumn{2}{c}{\textit{music 3}}     \\ \cline{2-7} 
\multicolumn{1}{c|}{}                                                                            & WER            & EER            & WER            & EER            & WER            & EER            \\ \hline
N/A                                                                                              & 19.34          & 38.88          & 13.62          & 29.82          & 18.92          & 33.70          \\
pre fine-tune                                                                                    & 18.99          & 36.89          & 13.26          & 28.92          & 18.40          & 32.25          \\
post fine-tune                                                                                   & \textbf{18.51} & \textbf{36.28} & \textbf{12.90} & \textbf{27.80} & \textbf{18.08} & \textbf{32.03} \\ \hline
\end{tabular}
}
\end{table}

\begin{table*}[t]
\caption{Domain adaptation performance on music, chatbot domain with various prompt handling}
\label{table:post_finetune}
\begin{tabular}{c|cccccc|cccccc}
\hline
\multirow{2}{*}{Domain} & \multicolumn{6}{c|}{Music}                                                                                                                    & \multicolumn{6}{c}{Chatbot}                                                                                                              \\ \cline{2-13} 
                        & \multicolumn{2}{c|}{\textit{music 1}}                         & \multicolumn{2}{c|}{\textit{music 2}}                         & \multicolumn{2}{c|}{\textit{music 3}}    & \multicolumn{2}{c|}{\textit{org.}}                           & \multicolumn{2}{c|}{\textit{tech.}}                         & \multicolumn{2}{c}{\textit{food}}      \\ \hline
Text adaptation         & WER            & \multicolumn{1}{c|}{EER}            & WER            & \multicolumn{1}{c|}{EER}            & WER            & EER            & WER           & \multicolumn{1}{c|}{EER}            & WER           & \multicolumn{1}{c|}{EER}           & WER           & EER           \\ \hline
N/A (base)                     & 19.34          & \multicolumn{1}{c|}{38.88}          & 13.42          & \multicolumn{1}{c|}{29.62}          & 18.92          & 33.70          & 5.54          & \multicolumn{1}{c|}{21.80}           & 4.82          & \multicolumn{1}{c|}{6.69}          & 2.23          & 4.98          \\
\textsl{no prompt}               & 18.51          & \multicolumn{1}{c|}{36.28}          & 12.90          & \multicolumn{1}{c|}{28.10}          & 18.08          & 32.03          & 5.52          & \multicolumn{1}{c|}{20.87}          & 4.67          & \multicolumn{1}{c|}{5.93}          & 2.15          & 4.36          \\
\textsl{empty prompt}            & 18.72          & \multicolumn{1}{c|}{36.43}          & 13.02          & \multicolumn{1}{c|}{28.14}          & 18.02          & 31.86          & 5.55          & \multicolumn{1}{c|}{21.03}          & 4.70          & \multicolumn{1}{c|}{6.23}          & 2.19          & 4.67          \\
\textsl{upsample and mask}       & 18.86          & \multicolumn{1}{c|}{36.60}          & 13.16          & \multicolumn{1}{c|}{28.42}          & 17.98          & 31.86          & 5.48          & \multicolumn{1}{c|}{20.87}          & 4.72          & \multicolumn{1}{c|}{5.93}          & 2.16          & 4.82          \\
\textsl{soft prompt (d=30)}      & \textbf{18.27} & \multicolumn{1}{c|}{\textbf{35.90}} & 12.65          & \multicolumn{1}{c|}{27.60}          & \textbf{17.69} & \textbf{30.92} & 5.42          & \multicolumn{1}{c|}{20.38}          & 4.60          & \multicolumn{1}{c|}{\textbf{5.63}} & 2.06          & 4.20 \\
\textsl{soft prompt (d=50)}      & 18.39          & \multicolumn{1}{c|}{36.28}          & \textbf{12.34} & \multicolumn{1}{c|}{\textbf{27.03}} & 17.88          & 31.62          & \textbf{5.34} & \multicolumn{1}{c|}{\textbf{20.22}} & \textbf{4.56} & \multicolumn{1}{c|}{\textbf{5.63}} & \textbf{2.04} & \textbf{4.05} \\ \hline
\end{tabular}
\end{table*}

\begin{table*}[t]
\caption{Domain adaptation performance on music, chatbot domain with LM fusion}
\label{table:post_finetune_lm_fusion}
\begin{tabular}{c|cccccc|cccccc}
\hline
\multirow{2}{*}{Domain} & \multicolumn{6}{c|}{Music}                                                                                                                    & \multicolumn{6}{c}{Chatbot}                                                                                                              \\ \cline{2-13} 
                        & \multicolumn{2}{c|}{\textit{music 1}}                         & \multicolumn{2}{c|}{\textit{music 2}}                         & \multicolumn{2}{c|}{\textit{music 3}}    & \multicolumn{2}{c|}{\textit{org.}}                           & \multicolumn{2}{c|}{\textit{tech.}}                         & \multicolumn{2}{c}{\textit{food}}      \\ \hline
Test adaptation         & WER            & \multicolumn{1}{c|}{EER}            & WER            & \multicolumn{1}{c|}{EER}            & WER            & EER            & WER           & \multicolumn{1}{c|}{EER}            & WER           & \multicolumn{1}{c|}{EER}           & WER           & EER           \\ \hline
N/A (base)              & 19.34          & \multicolumn{1}{c|}{38.88}          & 13.42          & \multicolumn{1}{c|}{29.62}          & 18.92          & 33.70          & 5.54          & \multicolumn{1}{c|}{21.80}          & 4.82          & \multicolumn{1}{c|}{6.69}          & 2.23          & 4.98          \\
base + LM               & 17.99          & \multicolumn{1}{c|}{35.76}          & 12.32          & \multicolumn{1}{c|}{27.46}          & 17.66          & 30.62          & 5.45          & \multicolumn{1}{c|}{20.22}          & 4.77          & \multicolumn{1}{c|}{5.93}          & 2.12          & 4.36          \\
\textsl{soft prompt}             & 18.27          & \multicolumn{1}{c|}{35.90}          & 12.65          & \multicolumn{1}{c|}{27.60}          & 17.69          & 30.92          & \textbf{5.34} & \multicolumn{1}{c|}{20.22}          & 4.56          & \multicolumn{1}{c|}{5.63}          & \textbf{2.04} & \textbf{4.05} \\
\textsl{soft prompt} + LM        & \textbf{17.58} & \multicolumn{1}{c|}{\textbf{35.52}} & \textbf{12.10} & \multicolumn{1}{c|}{\textbf{26.80}} & \textbf{17.36} & \textbf{30.49} & 5.36 & \multicolumn{1}{c|}{\textbf{20.16}} & \textbf{4.49} & \multicolumn{1}{c|}{\textbf{5.34}} & 2.08 & \textbf{4.05} \\ \hline
\end{tabular}
\end{table*}

\subsection{Setups}

We conduct domain adaptation with text-only corpora on two target domains, \textit{music} and \textit{chatbot}.

\textbf{Baseline Model.}  The baseline model we employ is the ASR model proposed in \cite{FatWuLak2023}. Since the proposed method is not only applicable on LLaMA based ASR model, we refer tohe baseline architecture as speech-LLM in the following content. The speech-LLM model comprises 700M parameters and is trained on text-audio paired data from the video domain. The audio encoder is a CTC model with 24 Conformer layers \cite{GulQinCi2020} pretrained on video datasets. The embedding and decoder are from a pretrained LLM with 500M parameters. The LLM contains 8 LLaMA decoder layers and is pretrained on the same datasets used in training the 7B LLaMA base model \cite{TouLavIza2023}.

\textbf{Fine-tuning Strategies.  } We adopted various fine-tuning strategies as discussed in Section \ref{sec:soft_prompt}. Regarding the timing of fine-tuning,
\textsl{pre fine-tune} means we fine-tune the pre-trained LLaMA on the target domain text corpus prior to the speech-LLM training; \textsl{post fine-tune} represents fine-tuning the embedding and decoder of the speech-LLM on the text corpus after the speech-LLM has been trained.
In terms of prompt handling in post fine-tuning, we either fine-tune with \textsl{no prompt} or we adopt:
\begin{itemize}
\item \textsl{empty prompt} where we simply prepend \texttt{[INST] [/INST]} tokens to the input text tokens;

\item \textsl{upsample and mask}: it refers to the upsampling and masking text embedding techniques used in \cite{SaiPraBap2023}, where we randomly upsample each token embedding 1 to 2 times and randomly mask 50\% of the upsampled embeddings to construct pseudo audio embeddings;

\item \textsl{soft prompt}: method refers to our proposed two-step soft prompt fine-tuning method, where we experimented with a soft prompt of length 30 or 50.
\end{itemize}

\textbf{External LM Fusion. }In addition to the fine-tuning approaches, we also explored the external LM fusion approach on the speech-LLM ASR. We trained domain-specific external LMs using the text-only corpus from the target domains. We then interpolated the external LM score with the decoder output score on each candidate token during the beam search. Each external LM comprises 50M parameters with 6 LSTM layers.

\begin{table}[t]
\caption{$p$-value of significance test on soft prompt v.s. others}
\label{table:sig_test}
\resizebox{\columnwidth}{!}{
\begin{tabular}{l|ll|ll}
\hline
\multicolumn{1}{c|}{\multirow{2}{*}{Comparison pairs}} & \multicolumn{2}{c|}{Music} & \multicolumn{2}{c}{Chatbot} \\ \cline{2-5} 
\multicolumn{1}{c|}{}                                  & word  & entity & word  & entity  \\ \hline
\textsl{soft prompt} v.s. base                                   & 0.040       & 0.028        & 0.048       & 0.024         \\
\textsl{soft prompt} v.s. \textsl{no prompt}                              & 0.058       & 0.042        & 0.068       & 0.036         \\ \hline
\end{tabular}
}
\end{table}

\subsection{Results}

 We first conduct experiments to explore the effects of pre fine-tuning and post fine-tuning. The impact of each setting is demonstrated on the music domain adaptation task. The results are presented in Table \ref{table:pre_finetune}. For a fair comparison, we adopted \textsl{no prompt} in both pre and post fine-tuning. Given that the music domain is entity-heavy, we reported both the Word Error Rate (WER) and the Entity Error Rate (EER), with the EER calculated at the music entity level. As anticipated, post fine-tuning significantly outperforms pre fine-tuning.

We then explored the impact of various prompt handling approaches. We conducted domain adaptation tasks on the music and chatbot domains respectively. The results are presented in Table \ref{table:post_finetune}.
Given that both domains are entity-heavy, we report WER and EER for comparison. In terms of prompt handling, fine-tuning with \textsl{no prompt} performs marginally better than using an \textsl{empty prompt}, indicating that improper handling of prompts can undermine the effectiveness of domain knowledge fine-tuning.
The pseudo audio prompt, constructed from upsampling and random masking text embeddings, simplifies the generation task during fine-tuning, performance shows it injects less effective knowledge into the decoder. \textsl{soft prompt} fine-tuning performs better than other prompt handling methods when the length of soft prompts is appropriately chosen. For example, compared with music domain, the model prefers the longer prompt for chatbot domain since the utterances are usually longer than those of the music domain. We also trained soft prompts longer than 50 but observed no significant performance change.
Overall, compared with the baseline speech-LLM, text adaptation with the proposed \textsl{soft prompt} approach achieves a relative 5.5-8.0\% WER reduction and 6.2-9.4\% EER reduction on the music domain. It also achieves a relative 3.7-8.9\% WER reduction and 7.3-18.0\% EER reduction.
Compared with text adaptation fine-tuning with \textsl{no prompt}, the proposed method achieves an additional up to 6.0\% WER reduction and up to 9.6\% EER reduction relatively. 

We further conducted an experiment on text adaptation through fusion with an external domain specific LM. The LM fusion was applied to both the baseline speech-LLM and the model fine-tuned with the proposed \textsl{soft prompt} approach. Results are shown in Table \ref{table:post_finetune_lm_fusion}. We observed additional improvements on the soft prompt fine-tuned model with LM fusion, where a relative reduction of up to 3.9\% in WER and up to 2.8\% in EER was achieved on the music domain. Similarly, a relative reduction of up to 2\% in WER and up to 5.4\% in EER was achieved on the chatbot domain. These results indicate that the effects of LM fusion and the proposed approach complement each other, leading to enhanced performance in both domains.

\subsection{Analysis and Discussion}

To verify whether the improvement from the proposed approach occurred by chance, we conducted a significance test using bootstrap. We combined all test splits in each domain to verify domain-wise significance. Our test is similar to the bootstrap test in \cite{BisNey2004}.
For each domain, we randomly drew 1k examples with replacement as a bootstrap sample and repeated the drawing 1k times. We measured the error differences at both the word and entity levels, then reported the $p$-value in Table \ref{table:sig_test} with a significance level at $0.05$.
Compared with the base ASR, results show that \textsl{soft prompt} methods perform significantly better on both domains. Compared with \textsl{no prompt} text adaptation, the word-level improvement of using \textsl{soft prompt} is marginally above the significance threshold, but the entity-level improvement is quite significant.
These results showcase the effectiveness of the proposed prompt handling approach. The substantial improvements in domain entity recognition indicate the generation of more semantically reliable transcriptions.

\section{Conclusion}

In this study, we introduced a two-step soft prompt fine-tuning approach to facilitate text adaptation in LLM-based ASR systems. Our results highlight the critical role of prompt management and demonstrate the efficacy of our proposed methods. The domain-specific soft prompt designed in this approach is uniformly applied across the domain during the fine-tuning process. Moving forward, we will delve deeper into the learning of pseudo audio embeddings for individual text sentences to further improve text adaptation effectiveness.

\footnotesize
\bibliographystyle{IEEEbib}
\bibliography{mybib}

\end{document}